\documentclass[twoside]{article}

\usepackage[accepted]{aistats2024}

\usepackage[table]{xcolor} 
\usepackage{hyperref}
\usepackage{url}
\usepackage{amsthm}
\usepackage{graphicx}
\usepackage{mathtools}
\usepackage{amsmath}
\usepackage{amsthm}
\usepackage{bbm}
\usepackage{mathrsfs} 
\usepackage{amsfonts}
\usepackage{amssymb}
\usepackage{comment}
\usepackage{algorithmic}
\usepackage{algorithm}
\usepackage{multirow}
 \usepackage[normalem]{ulem}
 \useunder{\uline}{\ul}{}
\usepackage{tabularx}
\usepackage{multirow}
\usepackage{makecell}
\usepackage{booktabs}

\usepackage{placeins}

\usepackage{caption}
\usepackage{subcaption}
\usepackage{wrapfig}

\DeclareMathOperator{\Tr}{Tr}

\newcommand{\norm}[1]{ \left\| #1 \right\| }
\newcommand{\prob}[2]{\mathbb{P}_{#1} \left[ #2 \right]}
\DeclareMathOperator*{\argmax}{arg\,max}
\DeclareMathOperator*{\argmin}{arg\,min}
\DeclareMathOperator*{\E}{\mathbb{E}}

\newcommand{\D}{\mathcal{D}}

\renewcommand{\t}[1]{{#1}^\top}

\newcommand{\argmins}{\argmin_{S\subseteq V, |S| \leq n_s}}

\newcommand{\argmaxs}{\argmax_{S\subseteq V, |S| \leq n_s}}
\newcommand{\xl}{x_{\mathcal{L}}}
\newcommand{\xv}{x_{\mathcal{V}}}
\newcommand{\muxv}{\mu_{\xv}}
\newcommand{\muxl}{\mu_{\xl}}
\newcommand{\noise}{\epsilon}
\newcommand{\noisel}{\noise_{\mathcal{L}}}
\newcommand{\noisev}{\noise_{\mathcal{V}}}
\newcommand{\fl}{f_{\mathcal{L}}}
\newcommand{\fv}{f_{\mathcal{V}}}

\newcommand{\flp}{f^p_{\mathcal{L}}}
\newcommand{\fvp}{f^p_{\mathcal{V}}}
\newcommand{\Fl}{F_{\mathcal{L}}}
\newcommand{\Fv}{F_{\mathcal{V}}}
\newcommand{\Fls}{F^S_{\mathcal{L}}}
\newcommand{\Fvs}{F^S_{\mathcal{V}}}

\newcommand{\Dl}{\mathcal{X}_\mathcal{L}}
\newcommand{\Dv}{\mathcal{X}_\mathcal{V}}

\newcommand{\dl}{d_{\mathcal{L}}}
\newcommand{\dv}{d_{\mathcal{V}}}
\newcommand{\Ul}{T_{\mathcal{L}}}
\newcommand{\Uv}{T_{\mathcal{V}}}
\newcommand{\Cds}{C^S_{\D}}
\newcommand{\Cdv}{C^V_{\D}}

\newcommand{\uf}{u} 
\newcommand{\ufnv}{\overline{\uf}_\mathcal{V}}
\newcommand{\ufnl}{\overline{\uf}_\mathcal{L}}
\newcommand{\U}{\mathcal{U}} 

\newcommand{\Cnus}{\overline{C^S_\U}}
\newcommand{\Cnuv}{\overline{C^V_\U}}

\newcommand{\Ftheory}{F_{\text{cov}}(S)}
\newcommand{\Ffinal}{F_{\text{ClipCov}}(S)}
\newcommand{\Fcross}{F_{\text{class}}(S)}
\newcommand{\Fself}{F_{\text{self}}(S)}
\newcommand{\Fdiv}{F_{\text{inter}}(S)}
\newcommand{\Fcrossreg}{F_{\text{class}}^{\text{reg}}(S)}

\newcommand{\Flabelsim}{F_{\text{label}}(S)}
\newcommand{\Dyy}{\mathcal{Y}}
\newcommand{\Dy}{\D_{\Dyy}}

\newcommand{\errzs}[1]{\mathcal{E}_{zs}(#1)}

\newcommand{\cross}[1]{{\text{sim}(#1)}}

\newcommand{\mmcl}{\text{CLIP}}

\newcommand{\method}{\textsc{ClipCov}}
\newcommand{\ssize}{n_s}

\theoremstyle{plain}
\newtheorem{theorem}{Theorem}[section]

\theoremstyle{definition}
\newtheorem{definition}[theorem]{Definition}

\theoremstyle{remark}

\begin{document}

%
\runningtitle{Data-Efficient Contrastive Language-Image Pretraining}

%

\twocolumn[

\aistatstitle{Data-Efficient Contrastive Language-Image Pretraining: \\Prioritizing Data Quality over Quantity}

\aistatsauthor{ Siddharth Joshi \And Arnav Jain \And  Ali Payani \And Baharan Mirzasoleiman }

\aistatsaddress{UCLA CS \And  UCLA CS \And Cisco Systems Inc. \And UCLA CS } ]

\begin{abstract}
Contrastive Language-Image Pre-training (CLIP) on large-scale image-caption datasets learns representations that can achieve remarkable zero-shot generalization. However, such models require a massive amount of pre-training data. Improving the quality of the pre-training data has been shown to be much more effective in improving CLIP's performance than increasing its volume. Nevertheless, finding small subsets of training data that provably generalize the best has remained an open question. In this work, we propose the first theoretically rigorous data selection method for CLIP. We show that subsets that closely preserve the cross-covariance of the images and captions of the full data provably achieve a superior generalization performance. Our extensive experiments on ConceptualCaptions3M and ConceptualCaptions12M demonstrate that subsets found by \method\ achieve over 2.7x and 1.4x the accuracy of the next best baseline on ImageNet and its shifted versions. Moreover, we show that our subsets obtain 1.5x the average accuracy across 11 downstream datasets, of the next best baseline.
The code is available at: \url{https://github.com/BigML-CS-UCLA/clipcov-data-efficient-clip}. 
\end{abstract}
 \vspace{-2mm}
\section{INTRODUCTION}
Contrastive Language-Image Pretraining (CLIP) has recently showed an impressive success, by  
enabling zero-shot recognition ability,  transferability to downstream tasks, and 
learning robust representations to distribution shift \cite{pmlr-v139-radford21a}. CLIP is trained on large image-caption datasets, by maximizing the alignment between paired image-captions representations and minimizing the agreement between image-caption representations of different pairs. Achieving this success, however, requires 1000 times larger datasets than traditional vision datasets like ImageNet \cite{imagenet_cvpr09}. 
For example, CLIP \cite{pmlr-v139-radford21a} and ALIGN \cite{jia2021scaling} are trained on 400M and 1B image-captions pairs crawled from the internet. 
This raises a key question of whether such a massive data is necessary to achieve superior performance and robustness. \looseness=-1

There have been recent efforts in answering this question. \cite{gadre2023datacomp} showed that smaller, more stringently filtered datasets can lead to models that generalize better than larger datasets coming from the same pool. For example, 1.4B images-caption pairs with highest similarity to ImageNet images, and with a high image-caption similarity outperform the full 12.8B data for zero-shot classification on ImageNet. While the effectiveness of such simple filtering strategies confirms the importance of the data quality for training CLIP, such strategies cannot further improve the data efficiency of language-image pretraining. In fact, it is not clear how one can select \textit{small subsets} of the training data that \textit{provably} 
generalize best, when trained on.

Finding small image-caption subsets with superior generalizability is indeed very challenging and demands fundamental understanding of the representation learning mechanism of CLIP. Indeed, the complex multimodal nature of CLIP makes existing data selection techniques 
inapplicable. Supervised data selection techniques that select examples based on per-example gradient \cite{pmlr-v202-yang23g, pooladzandi2022adaptive}, loss \cite{paul2023deep}, or entropy of the predictions \cite{coleman2020selection} are not applicable to CLIP. This is because the contrastive CLIP loss and its gradient depend on the entire dataset and excluding any example affects the gradient of all the examples. The recent data selection technique of \cite{pmlr-v202-joshi23b} for uni-modal contrastive learning, which finds images with central representations, also does not generalize to the multimodal scenario. Data selection for CLIP is inherently more complicated, due to the interaction between the image and text modalities. \looseness=-1

In this work, we address the above challenge for the first time. We rely on recent theoretical results of \cite{pmlr-v206-nakada23a} that showed that the CLIP representations are determined by the cross-covariance matrix of the image-caption data. We show that the subsets that 
closely capture
the cross-covariance of the image-caption pairs in the data can guarantee similar zero-shot 
generalization performance for CLIP.

{We confirm the effectiveness and scalability of our proposed technique through extensive experiments on the Conceptual Caption (CC) 3M \cite{sharma2018conceptual} and CC 12M \cite{cc12m} datasets. We show that our subsets (of sizes 5\% - 50\%) outperform equal size subsets found using several CLIP data filtering baselines, including CLIP score \cite{gadre2023datacomp}, C-RHO \cite{maini2023tmars}, SemDeDup \cite{abbas2023semdedup} and random selection. The subsets selected by \method\ achieve over  2.7x and 1.4x the accuracy of the next best baseline on ImageNet and its shifted versions \cite{imgnetobjectnet, deng2009imagenet, imgneta, imgnetr, imgnetv2, imgnetsketch}, respectively. Additionally, we demonstrate that our selected subsets obtain 1.5x the accuracy of the next best baseline, across 11 different downstream datasets.}




\section{RELATED WORK}
\textbf{Multimodal Contrastive Learning} 
Recently, Contrastive Language-Image Pre-training (CLIP) on large datasets comprised of paired images and captions has shown remarkable zero-shot generalization performance and transferability to a variety of downstream tasks. In particular, CLIP \cite{pmlr-v139-radford21a} and ALIGN \cite{jia2021scaling} trained on 400M/1B image-caption pairs achieve comparable accuracy to SOTA supervised learning across several tasks, without the need for any further training. Several recent studies aimed to improve the data-efficiency and performance of CLIP via data augmentation on image and text modalities \cite{li2021supervision, mu2022slip}, and imposing geometrically consistency in the image and text space \cite{goel2022cyclip}. \looseness=-1

\textbf{Multimodal Contrastive Learning Theory} 
 A few recent works have studied dynamics of multimodal contrastive learning. \cite{pmlr-v202-zhang23an} extends the results of \cite{haochen2022provable}, which showed the equivalence between the matrix factorization objective and the spectral contrastive loss, to the spectral multimodal contrastive loss. \cite{pmlr-v206-nakada23a} showed that for linear models, 
 each step of loss minimization by gradient descent can be seen as performing SVD on a contrastive cross-covariance matrix. We utilize the theory of \cite{pmlr-v206-nakada23a} to characterize the subsets that contribute the most to MMCL and guarantee superior generalization performance for CLIP. 


\textbf{Data Filtering for Multi-Modal Contrastive Learning}
Large image-caption datasets crawled from the internet often contain image-caption pairs that are uninformative, or contain unreliable or wrong captions. Hence, such datasets are often filtered before being used for training. Several data filtering methods have been proposed recently. {Such methods often use a pre-trained CLIP model to filter examples based on their similarity of image-caption representations \cite{gadre2023datacomp,maini2023tmars}.} 
Some other methods \cite{maini2023tmars, devil} address dataset-specific problems, e.g. the presence of text in large number of images, and do not yield useful subsets on other datasets.
While data filtering methods are essential to filter potentially wrong or irrelevant examples, they cannot find generalizable subsets from filtered datasets. Another recently proposed method aims to reduce the redundancy in the dataset by eliminating examples with similar image representations \cite{abbas2023semdedup}. {However, this method too drastically fails to find the most generalizable subsets from large image-caption datasets, as we will confirm experimentally.}  


\textbf{Data Selection for Supervised and Self-supervised Learning} Data efficiency in supervised learning has been the subject of extensive research, as evidenced by a long line of work \cite{coleman2020selection, mindermann2022prioritized, pooladzandi2022adaptive, pmlr-v202-yang23g}. However, applying these techniques directly to MMCL is not possible due to the absence of labels. Additionally, loss-based methods may not be suitable for MMCL because the loss of examples in MMCL depends on all examples in a batch. While data-efficiency for uni-modal contrastive learning has been studied before in \cite{pmlr-v202-joshi23b}, it cannot be transferred to multimodal learning due to the fundamental differences in data i.e. no augmentations and paired data from different modalities. 


\vspace{-1mm}
\section{PROBLEM FORMULATION}
\textbf{Data Distribution}
Let $\mathcal{D}=\{(\xv^i, \xl^i)\}_{i\in V}$ be a set of $n=|V|$ image-caption pairs i.e. the full training data available to us, drawn from $K$ latent classes i.e. $V = \cup_{k \in [K]} V_k$, where $\xv^i$ denotes the image and $\xl^i$ denotes the caption of the $i$-th example. Moreover, let $\Dv$ be the set of images and $\Dl$ be the set of captions in $\D$. To model the notion that paired image-captions describe the same underlying object, let image-caption pair $(\xv^i, \xl^i) \in \D$ be generated as follows:
\begin{align}
&\xv^i = \Uv (\uf^i + \noisev) \quad\quad \xl^i = \Ul (\uf^i + \noisel),
\end{align} where $\uf^i \in \mathbb{R}^d$ is the shared underlying feature vector for example $i$; $\Uv: \mathbb{R}^d \rightarrow \mathbb{R}^{\dv}$ and  $\Ul: \mathbb{R}^d \rightarrow \mathbb{R}^{\dl}$ are the mappings from underlying feature space to the vision and language data spaces; and $\noisev, \noisel$ are the noise in underlying features for vision and language, respectively. We refer to $\ufnv^i = \uf^i + \noisev^i$ and $\ufnl^i = \uf^i + \noisel^i$ as the noisy underlying feature for the image and caption of example $i$. The underlying feature $\uf^i$, for each image-caption pair is sampled independently of other pairs and of the noise $\noisev, \noisel$. Additionally, we assume $\forall i, \norm{\ufnv}^i, \norm{\ufnl}^i$ and $\norm{\Uv}, \norm{\Ul}$ is $\leq 1$. 



The shared underlying feature helps us capture the notion that paired image-captions represent the same underlying object (feature) e.g. `\texttt{a dog}'. 
The noise in the data distribution allows us to model both the occurrence of mismatched pairs e.g. an image of `\texttt{a dog}' matched with caption `\texttt{a cat}' as well as noise in data space for both images and texts e.g. an image of `\texttt{a dog with a cat in the background}' paired the caption `\texttt{a dog}' or the caption `\texttt{a dog with a cat}' paired with an image of `\texttt{a dog}'.

\textbf{Contrastive Language-Image Pre-training (CLIP)} CLIP is composed of a vision encoder $\fv: \mathbb{R}^{\dv} \rightarrow \mathbb{R}^r$ and a language encoder $\fl: \mathbb{R}^{\dl} \rightarrow \mathbb{R}^r$ that map input data in vision and language data space into a shared $r$-dimensional representation space, respectively. The vision and language encoders are trained by maximizing the representation similarity of paired image-captions and minimizing that of unpaired image-captions in every mini-batch, using the following multimodal contrastive loss:
\begin{align}\label{eq:clip_loss}
    \!\!&\mathcal{L}_{\text{CLIP}}(\fv, \fl) \!=\! \nonumber \\
    -\!\!&\E_{\xv, \xl \sim \D} \!\log \frac{\exp({\fv(\xv)}^\top \fl(\xl))}{\E_{\xl^- \sim \mathcal{X}_{\mathcal{L}}} \exp({\fv(\xv)}^\top \fl(\xl^-))} \nonumber \\
   -\!\! &\E_{\xv, \xl \sim \D} \!\log \frac{\exp({\fv(\xv)}^\top \fl(\xl)}{\E_{\xv^- \sim \mathcal{X}_{\mathcal{V}}} \exp({\fv(\xv^-)}^\top \fl(\xl))}.
\end{align}
For simplicity of theoretical analysis, we consider linear encoders 
where $\fv(\xv) = \Fv \cdot \xv$ and $\fl(\xl) = \Fl \cdot \xl$ where $\Fv \in \mathbb{R}^{r \times \dv}$ and $\Fl \in \mathbb{R}^{r \times \dl}$, used widely across machine learning literature \cite{pmlr-v206-nakada23a, ji2021power, pmlr-v202-xue23d}. Additionally, we use the linear multimodal contrastive loss used in \cite{pmlr-v206-nakada23a}:
\begin{align}\label{eq:lin_loss}
    \mathcal{L}(\Fv, \Fl)& = -\frac{1}{2n(n-1)}\sum_{i \in V}\sum_{\substack{j \in V \\ j\neq i}}(A_{ij} - A_{ii})\\ - 
    &\frac{1}{2n(n-1)} \sum_{i \in V}\sum_{\substack{j \in V \\ j\neq i}}(A_{ji}\!-\!A_{ii}) + \frac{\rho}{2}\| \Fv^\top \Fl \|_F^2, \nonumber
\end{align}
where $A_{ij}\coloneqq {(\Fv \xv^i)}^\top (\Fl \xl^j)$. \cite{pmlr-v206-nakada23a} shows that both the CLIP loss and the linear multimodal contrastive loss can be derived from a generalized form of the multimodal contrastive loss i.e. aliging representations of paired image-captions and separating representations of unpaired image-captions.


Note that we only use linear encoders and the linear multi-modal contrastive loss function in our theoretical analysis; the experiments in Section \ref{sec:experiments} are conducted with non-linear encoders and the CLIP loss in Eq. \eqref{eq:clip_loss}.

\textbf{Zero-Shot Classification} After training, the model is evaluated via zero-shot classification on different downstream image classification tasks. A downstream task $\Dy$ is defined as a classification task on unseen data from a set of $\Dyy$ classes.
For zero-shot classification on downstream task $\Dy$, we use the language encoder $f_\mathcal{L}$ to encode the label 
of each class $y \in \Dyy$;
using a set of pre-engineered templates, e.g. `\texttt{A photo of a \{label\}}' to create several captions representing `\texttt{\{label\}}' \cite{pmlr-v139-radford21a}.
Then, the classification of an example $\xv$ is $zs_{f_V, f_L}(\xv) = \argmax_{k \in \Dy} \frac{\fv{\xv} \cdot z_k}{\norm{\fv{\xv}}\norm{z_k}}$,
where $z_k = \E_{\xl s.t. y(\xl) = k}[\fl(\xl)]$ is the average representation of templates obtained using $\fl$ for class label $k$. That is an example $\xv$ is classified by the closest (average) template representation. 
The zero-shot error of $\fv, \fl$ is defined as the fraction of misclassified examples using the trained vision and language encoders $f_V, f_L$: \looseness=-1
\begin{align}
\errzs{\fv, \fl} := \prob{\xv \sim \Dy}{y(\xv) \neq zs_{f_V, f_L}(\xv)}.
\end{align}




\textbf{Finding Generalizable Multimodal Subsets}
Our goal is to find a subset of training image-caption data $S \subseteq V$ of size at most $\ssize\geq|S|$, such that encoders trained on the subset achieve similar generalization, across downstream tasks using zero-shot evaluation, to encoders trained on the full training data $V$. To do so, we formulate the problem as finding a subset $S$ such that the encoders learnt on the subset closely approximate the encoders learnt on the full training data $V$:
\begin{align}\label{eq:problem_formulation}
   S^* = \argmins \norm{\Fvs - \Fv} + \norm{\Fls - \Fl}
\end{align}
where $\Fvs, \Fls$ are the vision and language encoders learnt on the subset $S$ and $\Fv, \Fl$ are the encoders learnt on the the full training data $V$. 







\vspace{-2mm}\section{FINDING THE MOST GENERALIZABLE SUBSETS} \label{sec:method}

In this section, we first theoretically characterize how well the encoders learnt on an arbitrary subset $S$ approximate the encoders learnt on the full (training) data $V$. Then, we present \method, our algorithm for efficiently finding $S^*$, the most generalizable subset, from a massive corpus of image-caption pairs. 

To do so, we rely on the recent theoretical results showing that the training dynamics on the full data $V$ are determined by the cross-covariance matrix of all the image-caption pairs in the dataset \cite{pmlr-v206-nakada23a}. 
The centered cross-covariance matrix of the full data $\Cdv$ is defined as follows:
\begin{align}\label{eq:cross_cov_mtx}
    \Cdv := \frac{1}{|V|}\sum_{i \in V} (\xv^i - \muxv) {(\xl^i - \muxl)}^\top, 
\end{align}
where $\muxv = \E_{\xv \in \Dv} \xv$ is the center of vision data and $\muxl = \E_{\xl \in \Dl} \xl$ is the center of language data. \footnote{Since $|V|$ is large, we replace $|V| - 1$ with $|V|$ for simplicity.} The cross-covariance matrix for image-caption data captures the covariance between paired image-captions. 

The linear loss function in Eq. \eqref{eq:lin_loss} can be rewritten as the SVD objective function: 
\begin{align}\label{eq:lin_loss=svd}
     \mathcal{L}(\Fv, \Fl) = -\Tr(\Fv \Cdv \Fl^\top) + \frac{\rho}{2}\| \Fv^\top \Fl \|_F^2.
\end{align}

Likewise, dynamics of training on the subset is determined by the cross-covariance of the subset:
\begin{align}
    \mathcal{L}(\Fvs, \Fls) &=  -\Tr(\Fvs \Cds {\Fls}^\top) + \frac{\rho}{2}\| {\Fvs}^\top \Fls \|_F^2,
\end{align}
where $\Cds$ is the data cross-covariance matrix of the subset $S$. 

Hence, we see that if $\Cds$, the cross-covariance of the subset $S$, closely approximates $\Cdv$, the cross-covariance of the full data, by minimizing the contrastive multimodal loss, the encoders learnt on the subset $S$ will be similar to the encoders learnt on the full data $V$. 

\subsection{Preserving the Cross-Covariance of Data}\label{sec:method:theory}

To preserve the cross-covariance of the full data, we can preserve the cross-covariance of noisy image and caption underlying features. 
Let 
\begin{align}
\Cnuv = \frac{1}{|V|}\sum_{i \in V} (\ufnv^i - \E_{i \in V}{\ufnv^i}) {(\ufnl^i - \E_{i \in V}{\ufnl^i})}^\top, \nonumber \\ 
\Cnus = \frac{1}{|S|}\sum_{i \in S} (\ufnv^i - \E_{i \in S}{\ufnv^i}) {(\ufnl^i - \E_{i \in S}{\ufnl^i})}^\top, \nonumber
\end{align}
be the cross-covariance of noisy underlying features for the full data $V$ and subset $S$, respectively. Then we have that,




\vspace{-3mm}\begin{align}
    \norm{\Cdv - \Cds} &= \norm{\Uv \Cnuv \Ul^\top - \Uv \Cnus \Ul^\top} \\
    &\leq \norm{\Uv}\norm{\Ul}\norm{\Cnuv - \Cnus} \\
    &\leq \norm{\Cnuv - \Cnus},
\end{align}
where the last inequality holds since $\norm{\Uv} = \norm{\Ul} \leq 1$.

We see that if the subset $S$ preserves the cross-covariance of noisy underlying feature of $V$, then it also preserves the data cross-covariance of the full data.

To preserve the cross-covariance matrix of underlying features in Eq. \eqref{eq:cross_cov_mtx}, we need to (1) preserve the centers of images $\muxv$ and captions $\muxl$ in every latent class; 
and (2) preserve the cross-covariance of examples in every latent class by selecting image-caption pairs 
that are centrally located in different subpopulations of the latent class and represent its different subgroups. 

Next, we show how we can find a subset with the above two properties to closely preserve
the cross-covariance of the full data. 

\textbf{Preserving Centers of Latent Classes}

Let $\mu^{S_k}_\mathcal{V} = \E_{i \in S_k} \ufnv^i$ and $\mu^{S_k}_\mathcal{L} = \E_{i \in S_k} \ufnl^i$ be the centers of the noisy underlying features for images and captions in subset  $S_k$ selected from latent class $k$, respectively. Likewise, 
let $\mu^{V_k}_\mathcal{V} = \E_{i \in V_k} \ufnv^i$ and $\mu^{V_k}_\mathcal{L} = \E_{i \in V_k} \ufnl^i$ be the centers of the noisy underlying features for images and captions of latent class $k$ in full data, respectively.

We now bound the error in preserving the centers of images and captions: 
{\allowdisplaybreaks
\begin{align}
    &\norm{\mu^{S_k}_\mathcal{V} - \mu^{V_k}_\mathcal{V}} + \norm{\mu^{S_k}_\mathcal{L} - \mu^{V_k}_\mathcal{L}} \nonumber\\
    &\hspace{1cm}= \norm{\frac{1}{|S_k|}\sum_{i \in S_k} \ufnv^i - \frac{1}{|V_k|} \sum_{j \in V_k} \ufnv^j} \nonumber \\ 
    &\hspace{1.5cm}+ \norm{\frac{1}{|S_k|}\sum_{i \in S_k} \ufnl^i - \frac{1}{|V_k|}\sum_{j \in V_k} \ufnl^j}  \nonumber\\
    &\hspace{1cm}\leq \norm{\frac{1}{|S_k|}\sum_{i \in S_k} \ufnv^i - \frac{1}{|V_k|} \sum_{j \in V_k} \ufnl^j} \nonumber \\ 
    &\hspace{1.5cm}+ \norm{\frac{1}{|S_k|}\sum_{i \in S_k} \ufnl^i - \frac{1}{|V_k|}\sum_{j \in V_k} \ufnv^j}  \nonumber \\
    &\hspace{1.7cm}+ \underbrace{2 \norm{\mu^{V_k}_\mathcal{V} - \mu^{V_k}_\mathcal{L}}}_{\text{alignment of full data centers}}.
\end{align}
}
The alignment of centers of latent class $V_k$ in full data refers to the similarity of the image and caption centers of the noisy underlying features and is independent of the subset $S_k$. Moreover, with sufficiently small noise in underlying feature, full data class centers are nearly identical as the true underlying feature is shared across images and captions within a latent class. Hence we get:
\begin{align}
&\norm{\mu^{S_k}_\mathcal{V} - \mu^{V_k}_\mathcal{V}} + \norm{\mu^{S_k}_\mathcal{L} - \mu^{V_k}_\mathcal{L}}\nonumber\\
&\lessapprox \frac{1}{|S_k|} \frac{1}{|V_k|} \Bigg\|\sum_{\substack{i \in S_k \\ j \in V_k}} \ufnv^i - \ufnl^j\Bigg\| + \frac{1}{|S_k|} \frac{1}{|V_k|} \Bigg\|\sum_{\substack{i \in S_k  \\ j \in V_k}} \ufnl^i - \ufnv^j\Bigg\| \label{eq:preserve_center}
\end{align}

That is, we can minimize Eq. \eqref{eq:preserve_center} to find a subset that preserves image and caption centers, for each latent class $k$. 

However, in practice, we do not have access to these noisy underlying features. Instead, we approximate them using the representations of a vision and language encoder trained with the multimodal contrastive loss on the full data. We refer to these encoders as the proxy vision encoder $\fvp$ and the proxy language encoder $\flp$. Vision and language encoders trained on the full data recover the corresponding noisy underlying features, up to orthogonal transformation \cite{pmlr-v206-nakada23a}. Using the proxy encoders, we now introduce the notion of cross-modal similarity to help us minimize Eq. \eqref{eq:preserve_center}. \looseness=-1

\begin{definition}[\textbf{Cross-Modal Similarity}]\label{def:cross_modal_dist}
We define the cross-modal similarity between any two image-caption pairs $i, j \in V$ as the sum of the cosine similarities between the representations of the image of $i$ with caption of $j$ and caption of $i$ with image of $j$.
Formally, $\forall i, j \in V$, we have:
\begin{align}
    \cross{i, j} = \left<\fvp(\xv^i)^\top, \flp(\xl^j)\right> + \left<\fvp(\xv^j)^\top, \flp(\xl^i)\right>.\nonumber
\end{align}
\end{definition}

Since, the norm of underlying feature vectors $\leq 1$ and orthogonal transformations preserve inner products, we can minimize Eq. \eqref{eq:preserve_center} by maximizing the cross-modal similarity of the corresponding examples: $\sum_{\substack{i \in S_k \\ j \in V_k}} \cross{i, j}$. 


Thus, we can preserve centers for the full data by maximizing the following objective:
\begin{equation}
    \sum_{k \in [K]} \frac{1}{|V_k|}\sum_{\substack{i \in S_k \\ j \in V_k}}  \cross{i, j},
\end{equation}
where we normalize the objective for latent class $k$ by the size of the latent class $|V_k|$ to prevent larger classes from dominating the objective. 


In practice, the data within latent classes is often imbalanced.
To prevent large subgroups within latent classes from dominating the objective, we penalize the similarity between selected examples to encourage diversity in the subset. Thus, we preserve the centers and alignment of centers, while ensuring diversity of the selected examples, using the following objective: 
\begin{align}\label{eq:fcross} &
\Fcross \!:=\! \sum_{k \in K} \frac{1}{|V_k|} \Big( \sum_{\substack{i \in S_k \\ j \in V_k}} \cross{i, j} - \frac{1}{2}\sum_{\substack{i \in S_k \\ j \in S_k}} \cross{i, j} \Big).\nonumber
\end{align}

\vspace{-1mm}\textbf{Preserving Cross-covariance with CLIP Score}



Next, we aim to select a subset of examples that capture the cross-covariance between image-caption pairs within every latent class. To do so, we need to find examples that are centrally located in different subpopulations within every latent class and represent its different subgroups.
By minimizing the CLIP loss on such subsets and aligning their image-caption pairs, we also align other image-caption pairs in the corresponding subgroups within the latent class. 

To effectively find subsets that capture the covariance within every latent class, we use a pre-trained CLIP model to find examples that represent different subgroups of the data.
Having a pre-trained model with the CLIP loss, the above examples can be efficiently identified as pairs with largest cross-modal similarity between their own image and captions, using the following CLIP score objective: 
\vspace{-2mm}\begin{align}
\Fself = \sum_{i \in S} \cross{i, i}. \nonumber    \vspace{-4mm}
\end{align}
Examples found using the above CLIP score objective are similar to many other pairs in their latent class, and are most centrally located in different subgroups.

\begin{figure}[t]
    \centering
    \includegraphics[width=0.4\textwidth]{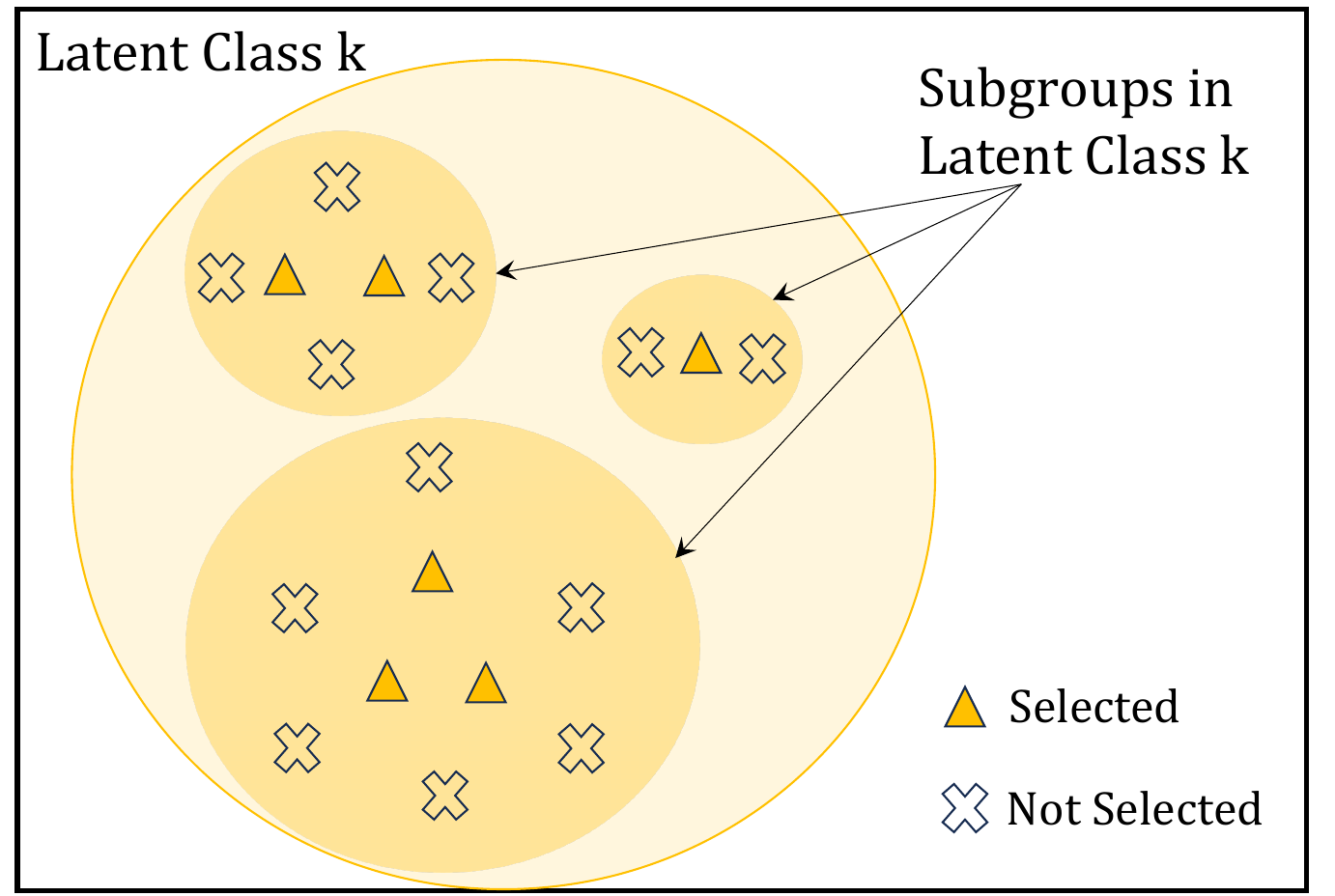}
    \caption{Visualization of examples selected by $\!\Fcross \!+\! \Fself\!$ in cross-modal similarity space. \method\ selects central examples that are representative of different subgroups in every latent class.}\label{fig:cross_diag}
    \vspace{-2mm}
\end{figure}

\textbf{CLIP score is effective to find larger subsets. }
Effectively, training with the CLIP loss aligns different image-caption pairs. In doing so, subgroups of image-caption pairs that have large cross-modal similarity to each other get close in the representation space. 
This is because groups of similar examples together introduce a large gradient during the training to pull their images and captions together. Hence, 
the most central example in each subgroup will have the largest cross modal similarity between its image and caption. Therefore, CLIP score can efficiently find examples that are centrally located in different subgroups of the training data. 
If the pre-training data is large and diverse, examples found by CLIP score obtain a superior performance on various downstream tasks, as they contain the most central examples in various subgroups of the data. Nevertheless, when the subset size is small, such examples cannot capture the center of latent classes accurately and $\Fcross$ is crucial to achieve superior generalization performance.

\begin{algorithm}[t]
\caption{\method}\label{alg}
\begin{algorithmic}[1]
\STATE \textbf{Input:} Dataset $V$, Subset size $\ssize$, proxy encoders: $\fvp$ and $\flp$ to calculate $\Ffinal$ 
\STATE \textbf{Output:} Subset $S$
\STATE $\{V_1, ..., V_K \} \leftarrow$ approximate latent classes
\STATE $S \leftarrow \{\}$
\STATE $F(S) := \Ffinal$ 
\hspace{2.6cm}
$\triangleright$ Eq. (\ref{eq:finaleq})
\STATE $S \leftarrow \emptyset \hspace{5cm}\triangleright$ \textbf{Greedy}
\WHILE{$|S| \leq \ssize$}
\STATE $e \leftarrow \argmax_{e \in V \setminus S} F(e|S)$
\STATE $S \leftarrow S \cup \{ e\}$
\ENDWHILE
\STATE {$S_1 \leftarrow \emptyset, S_2 \leftarrow S \hspace{2.11cm}\triangleright$ \textbf{Double Greedy}\looseness=-1 }
\FOR{$e \in S$}
\STATE $a \leftarrow F(e|S_1)$
\STATE $b \leftarrow F(S_2\setminus \{e\}) - F(S_2)$
\IF{$a\geq b$}
\STATE $S_1 \leftarrow S_1 \cup \{e\}$
\ELSE
\STATE $S_2\leftarrow S_2 \setminus \{e\}$
\ENDIF
\ENDFOR
\STATE \textbf{return} $S_1$ or equivalently $S_2$
\end{algorithmic}
\end{algorithm}
Hence, we can preserve the cross-covariance of the data by maximizing the following objective $\Ftheory$:
\begin{align}
    \Ftheory := \Fcross + \Fself. \nonumber
\end{align}\vspace{-5mm}

To illustrate what kinds of examples are selected by this objective, we provide a visualization in Fig. \ref{fig:cross_diag} which shows the selected examples are similar to all the examples in the latent class, even from smaller subgroups. From Fig. \ref{fig:cross_diag}, we can see how such a subset is representative of the latent class and thus can capture the cross-covariance within the latent class.

\vspace{-2mm}\subsection{
Deriving the Final Objective for Finding the Most Generalizable Subset
}\label{sec:method:practical}

We now discuss three practical considerations that often arise when learning from large vision-language datasets, and account for them in the final objective to find the most generalizable subsets. \looseness=-1

\textbf{Label Centrality for Zero-shot Classification} 
While preserving the cross-covariance within latent classes allows us to ensure that images in a given latent class can correctly be paired with their corresponding captions, zero-shot classification measures similarity of images representations to the text representations of the \textit{labels} of the latent classes. This is highly sensitive to the name of the label being similar to the captions of the corresponding latent class. To explicitly ensure that the selected captions are similar to the labels used, we introduce $\Flabelsim$:
\begin{align}
    \Flabelsim = \sum_{k \in [K]} \sum_{i\in S_k}& \alpha \t{\flp(\xl^i)} \flp(y_k) \nonumber\\  &- \sum_{i\in S_k} \alpha \frac{\t{\flp(\xl^i)}\flp(y_k)}{|V_k|}, \nonumber
\end{align}
where $\alpha$ is the ratio of average cross-modal similarity to the average similarity in text \footnote{Empirically, we find $\alpha \approx \frac{1}{2}$.}  Here, the second term prevents domination of classes with very good similarity to the label. This improves the zero-shot performance on various downstream datasets. 

\textbf{Dealing with Imbalanced Data} 
In practice, when the sizes of latent classes are extremely imbalanced i.e. some latent classes in the training data are much larger than others, this leads to $\Fcross$ for large latent classes dominating the objective. Hence, we further regularize $\Fcross$ to avoid only selecting examples from large latent classes by deducting the following regularization term from the objective. 
\begin{align}
    &\Fcrossreg = \sum_{k \in K} \frac{1}{|V_k|} \sum_{\substack{i \in S_k \\ j \in V_k}} \frac{\cross{i, j}}{|V_k|},\nonumber 
\end{align}
which is approximately the average sum of intra-class cross-modal similarity of the selected subset $S$.



\textbf{Penalizing Inter-class Similarity}
Empirically, we find that ensuring that the examples selected for different latent classes are dissimilar yields more distinguishable representations for latent classes, improving performance across various 
downstream tasks. 
Thus, we minimize the average similarity of examples to other latent classes. The following objective, $\Fdiv$, formalizes this:

\begin{align}
    \Fdiv :=~ - \!\!\sum_{\substack{k_1, k_2 \in [K] \\ k_1 \neq k_2}}\sum_{i \in S_k} \sum_{j \in V_{k_2}} \frac{\cross{i, j}}{|V_{k_2}|}, \nonumber
\end{align}
where 
$\sum_{i \in S_k} \sum_{j \in V_{k_2}} \frac{\cross{i, j}}{|V_{k_2}|}$ is the average cross-modal similarity of image-caption pair $i$ to image-caption pairs in $V_{k_2}$. In practice, we can compute this average cross-modal similarity efficiently, by first averaging the image-caption representations of latent class $k_2$ and then computing the cross-modal similarity between examples $i \in V_{k_1}$ and the average image-caption representations of $V_{k_2}$.

\textbf{Final Objective}
Hence, the final objective for finding the most generalizable subset $S^*$ is:
\begin{align}
    &S^* \in \argmaxs \Ffinal , \quad \text{where} \\
    &\Ffinal := \label{eq:finaleq}\\
    &\hspace{1.5cm}\Ftheory 
    +\Flabelsim - \Fcrossreg + \Fdiv .\nonumber
\end{align}

\subsection{\method: Efficiently Finding the Most Generalizable Subset}\label{sec:method:clip_core}\vspace{3mm}

\begin{figure*}[htbp]
    \centering
    \begin{subfigure}[b]{0.3\textwidth}
        \centering
        \includegraphics[width=\linewidth]{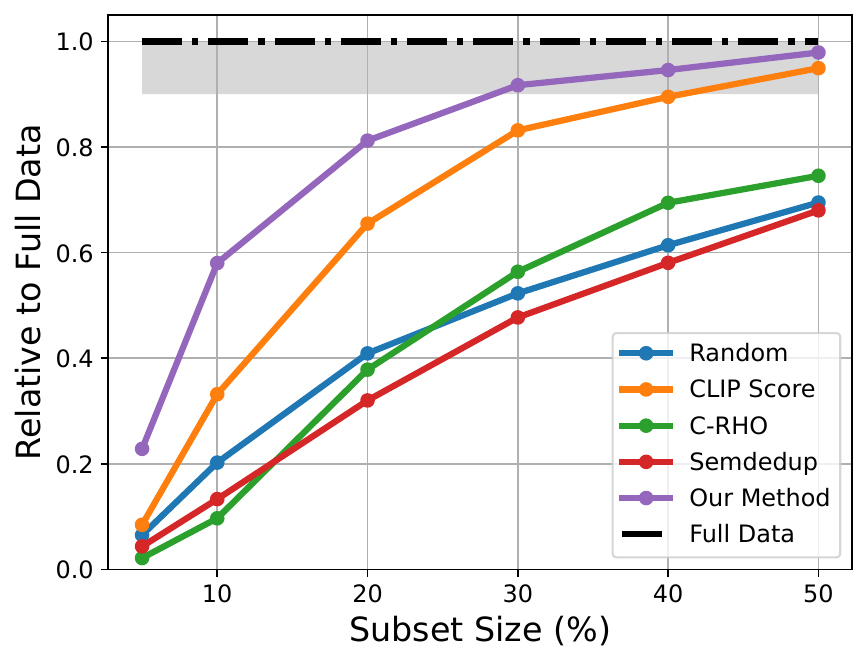}
        \caption{ImageNet}
        \label{fig:sub1}
    \end{subfigure}
    \hfill
    \begin{subfigure}[b]{0.3\textwidth}
        \centering
        \includegraphics[width=\linewidth]{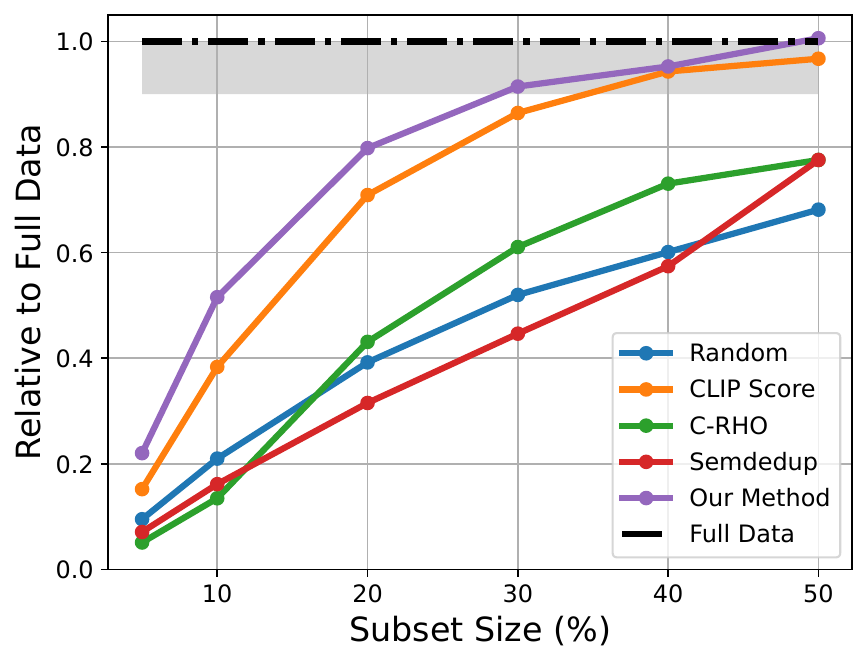}
        \caption{ImageNet Dist. Shift}
        \label{fig:sub2}
    \end{subfigure}
    \hfill
    \begin{subfigure}[b]{0.3\textwidth}
        \centering
        \includegraphics[width=\linewidth]{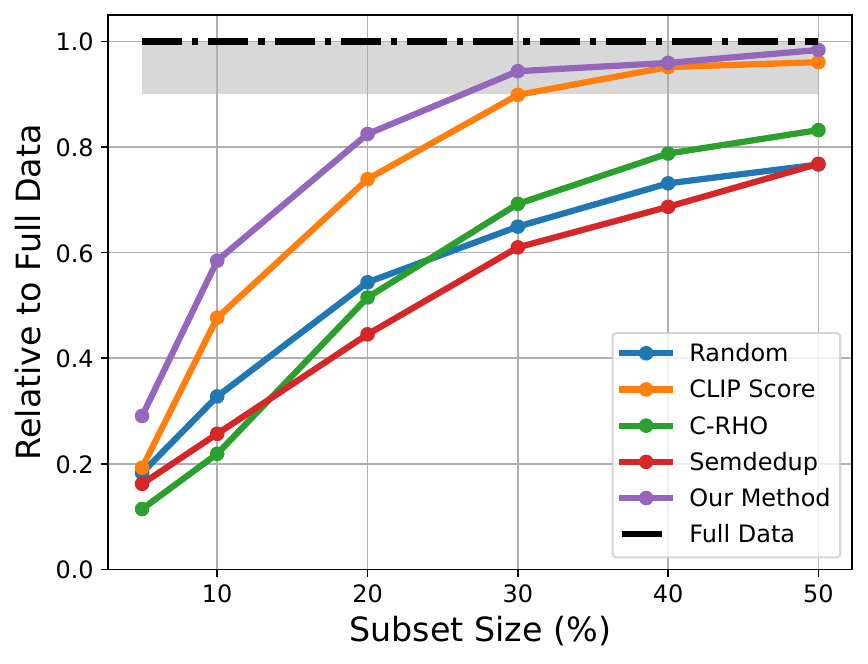}
        \caption{Avg. over 11 Datasets}
        \label{fig:sub3}
    \end{subfigure}
    \caption{ Performance across subset of different sizes selected from ConceptualCaptions3M. Gray region indicates accuracy within 90\% of that of full data.}
    \label{fig:main_results}
\end{figure*}

\begin{table*}[ht]
\centering
\caption{Performance of 5\% and 10\% subsets selected from ConceptualCaptions12M}\label{table:cc12m_results}
\begin{small}
\begin{tabular}{c|cccc}
\toprule
Subset Size & Method   & ImageNet & ImageNet Dist. Shift & Avg. over 11 Datasets \\   \midrule\midrule
\multirow{2}{*}{5\%}  & \cellcolor{gray!20}{CLIP Score}              & \cellcolor{gray!20}{5.10\%}     & \cellcolor{gray!20}{4.42\%}                  &\cellcolor{gray!20} 9.49\%   \\
                     & \textbf{\method}                & \textbf{13.61\%}   & \textbf{7.99\%}                 &  \textbf{11.68\%}  \\ \midrule
\multirow{2}{*}{10\%} 
                     & {CLIP Score}              & {11.02\%}     & {8.55\%}                  & {14.69\%}   \\
                     & \cellcolor{gray!20} \textbf{\method}                & \cellcolor{gray!20} \textbf{22.71\%}   & \cellcolor{gray!20} \textbf{12.76\%}                 & \cellcolor{gray!20} \textbf{16.87\%}  \\ \bottomrule
\end{tabular}
\end{small}
\end{table*}

\begin{table}[ht]
\centering
\begin{small}
\caption{Ablation over proxy encoders}\label{table:ablation_proxy}\vspace{-2mm}
\rowcolors{2}{gray!25}{white}
\begin{tabular}{c|ccc}
\toprule
Method   & ImageNet & ImgNet Shift& Avg.\\ \midrule\midrule
CLIP Score  & 5.01\%   & 3.16\%    & 7.35\%  \\
\textbf{\method}   & \textbf{6.70\%}   & \textbf{3.48\%}   & \textbf{9.10\%}  \\ \bottomrule
\end{tabular}
\end{small}
\vspace{-4mm}
\end{table}

\begin{table*}[ht]
\centering
\begin{small}
\caption{Ablation over objective for 10\% subset selected from ConceptualCaption3M}\label{table:ablation_obj}\vspace{-2mm}
\rowcolors{2}{gray!25}{white}
\begin{tabular}{c|ccc}
\toprule
Method   & ImageNet & ImageNet Dist. Shift & Avg. over 11 Datasets   \\ \midrule\midrule
$\Ftheory$   &  8.74\%&  5.17\%& 10.82\%\\
$\Ftheory + \Fdiv$   &  9.00\%   &  5.30\%    & 10.96\% \\
$\Ftheory + \Fdiv - \Fcrossreg$  & 8.94\%    & 5.10\%     & 11.29\%    \\
$\Ftheory + \Fdiv + \Flabelsim$  & 10.87\% & 5.73\%    & 11.27\%    \\
\textbf{\method}    & \textbf{11.33\%}   & \textbf{5.97\%}                 & \textbf{12.64\%} \\\bottomrule
\end{tabular}
\end{small}
\end{table*}
\begin{table*}[htbp]
\vspace{-1mm}
\centering
\caption{Performance across subset of different sizes selected from ConceptualCaptions3M}\label{table:main_results}\vspace{-2mm}
\begin{small}
\begin{tabular}{c|cccc}
\toprule
Subset Size & Method   & ImageNet & ImageNet Dist. Shift & Avg. over 11 Datasets \\  \midrule\midrule
\multirow{5}{*}{5\%} & Random    &  1.27\%                & 1.10\%                          & 3.95\%           \\
                     & \cellcolor{gray!20} C-RHO       &\cellcolor{gray!20} 0.42\%             &\cellcolor{gray!20} 0.59\%                          &\cellcolor{gray!20} 2.47\%            \\
                     & SemDeDup                        & 0.85\%             & 0.82\%                          & 3.50\%            \\
                     & \cellcolor{gray!20}\uline{CLIP Score}              & \cellcolor{gray!20}\uline{1.65\%}     & \cellcolor{gray!20}\uline{1.76\%}                  &\cellcolor{gray!20} \uline{4.16\%}   \\
                     & \textbf{\method}                & \textbf{4.46\%}   & \textbf{2.55\%}                 &  \textbf{6.28\%}  \\ \midrule
\multirow{5}{*}{10\%} & \cellcolor{gray!20} Random                          & \cellcolor{gray!20} 3.95\%             & \cellcolor{gray!20} 2.43\%                          & \cellcolor{gray!20} 7.08\%           \\
                     & C-RHO                           & 1.89\%             & 1.56\%                          & 4.73\%            \\
                     & \cellcolor{gray!20} SemDeDup                        & \cellcolor{gray!20} 2.60\%             & \cellcolor{gray!20} 1.87\%                          & \cellcolor{gray!20} 5.55\%            \\
                     & \uline{CLIP Score}              & \uline{6.48\%}     & \uline{4.44\%}                  & \uline{10.30\%}   \\
                     & \cellcolor{gray!20} \textbf{\method}                & \cellcolor{gray!20} \textbf{11.33\%}   & \cellcolor{gray!20} \textbf{5.97\%}                 & \cellcolor{gray!20} \textbf{12.64\%}  \\ \midrule
\multirow{5}{*}{20\%} & \cellcolor{gray!20} Random                          & \cellcolor{gray!20} 7.99\%             & \cellcolor{gray!20} 4.54\%                          & \cellcolor{gray!20} 11.75\%           \\
                     & C-RHO                           & 7.39\%             & 4.99\%                          & 11.13\%            \\
                     & \cellcolor{gray!20} SemDeDup                        & \cellcolor{gray!20} 6.25\%             & \cellcolor{gray!20} 3.65\%                          & \cellcolor{gray!20} 9.62\%            \\
                     & \uline{CLIP Score}              & \uline{12.79\%}     & \uline{8.21\%}                  & \uline{15.87\%}   \\
                     & \cellcolor{gray!20} \textbf{\method}                & \cellcolor{gray!20} \textbf{15.86\%}   & \cellcolor{gray!20} \textbf{9.24\%}                 & \cellcolor{gray!20} \textbf{17.82\%}  \\ \midrule
\multirow{5}{*}{30\%} & \cellcolor{gray!20} Random                          & \cellcolor{gray!20} 10.21\%             & \cellcolor{gray!20} 6.02\%                          & \cellcolor{gray!20} 14.03\%           \\
                     & C-RHO                           & 11.01\%             & 7.07\%                          & 14.96\%            \\
                     & \cellcolor{gray!20} SemDeDup                        & \cellcolor{gray!20} 9.32\%             & \cellcolor{gray!20} 5.17\%                          & \cellcolor{gray!20} 13.18\%            \\
                     & \uline{CLIP Score}              & \uline{16.24\%}     & \uline{10.01\%}                  & \uline{19.42\%}   \\
                     & \cellcolor{gray!20} \textbf{\method}                & \cellcolor{gray!20} \textbf{17.91\%}   & \cellcolor{gray!20} \textbf{10.59\%}                 & \cellcolor{gray!20} \textbf{20.39\%}  \\ \midrule
\multirow{5}{*}{40\%} & \cellcolor{gray!20} Random                          & \cellcolor{gray!20} 11.99\%             & \cellcolor{gray!20} 6.96\%                          & \cellcolor{gray!20} 15.80\%           \\
                     & C-RHO                           & 13.56\%             & 8.46\%                          & 17.02\%            \\
                     & \cellcolor{gray!20} SemDeDup                        & \cellcolor{gray!20} 11.34\%             & \cellcolor{gray!20} 6.65\%                          & \cellcolor{gray!20} 14.84\%            \\
                     & \uline{CLIP Score}              & \uline{17.48\%}     & \uline{10.92\%}                  & \uline{20.56\%}   \\
                     & \cellcolor{gray!20} \textbf{\method}                & \cellcolor{gray!20} \textbf{18.47\%}   & \cellcolor{gray!20} \textbf{11.03\%}                 & \cellcolor{gray!20} \textbf{20.73\%}  \\ \midrule
\multirow{5}{*}{50\%} & \cellcolor{gray!20} Random                          & \cellcolor{gray!20} 13.54\%& \cellcolor{gray!20} 7.89\%& \cellcolor{gray!20} 16.57\%\\
                     & C-RHO                           & 14.56\%& 8.98\%& 17.98\%\\
                     & \cellcolor{gray!20} SemDeDup                        & \cellcolor{gray!20} 13.28\%& \cellcolor{gray!20} 8.98\%& \cellcolor{gray!20} 16.60\%\\
                     & \uline{CLIP Score}              & \uline{18.54\%}& \uline{11.20\%}& \uline{20.76\%}\\
                     & \cellcolor{gray!20} \textbf{\method}                & \cellcolor{gray!20} \textbf{19.12\%}& \cellcolor{gray!20} \textbf{11.65\%}& \cellcolor{gray!20} \textbf{21.26\%}\\ \midrule
\end{tabular}
\end{small}
\vspace{-3mm}
\end{table*}


Here, we discuss how the proxy representations and latent classes required to solve Problem \eqref{eq:finaleq} are obtained. We then present \method\ and show how it can efficiently find this subset from massive datasets. 

\textbf{Obtaining Proxy Representations} We can use any pretrained CLIP as the proxy encoders to determine the proxy representations cross-covariance matrix. The effectiveness of \method\ is dependent on how closely the proxy representations recover the underlying features of the full data. Hence, we use the open-source pretrained CLIP encoders provided by \cite{pmlr-v139-radford21a}, which are trained on massive amounts of data and obtain impressive zero-shot generalization, thus are likely effectively recover the underlying features of the full data $V$.

\textbf{Approximating Latent Classes}
In practice, we do not have access to latent classes required to solve Problem \eqref{eq:finaleq}. Instead, we approximately recover latent classes via zero-shot classification using proxy \mmcl\ encoders. 
For zero-shot classification, we use 
1000 labels of ImageNet-1k. This allows finding fine-grained latent classes.

\textbf{Scaling to Massive Datasets}
Since $\Fdiv$ can be computed using the average representations of latent classes, in practice, \method\ only needs to compute pairwise cross-modal similarities within latent classes. Here, the fine-grained latent classes used also ensure that computing pairwise cross-modal similarities within latent classes is inexpensive.


{Maximizing Objective \eqref{eq:finaleq} is NP-hard as it requires evaluating an exponential number of subsets. To efficiently find a near-optimal subset, we note that $\Ftheory$ is non-monotone submodular, and $\Fdiv$, $\Flabelsim,$ $\Fcrossreg$ are modular. Hence, Objective \eqref{eq:finaleq} is non-monotone submodular. Thus, we can find a near-optimal subset using algorithms for non-monotone submodular function maximization under a cardinality constraint. To do so, we first use the greedy algorithm to find a subset, and then filter the subset by applying unconstrained submodular maximization \cite{mirzasoleiman2016fast}. 
The greedy algorithm starts with the empty set $S_0=\emptyset$, and at each iteration $t$, it chooses an element $e\in V$ that maximizes the marginal utility $F(e|S_{t})=F(S_{t}\cup\{e\}) - F(S_{t})$. Formally, $S_t = S_{t-1}\cup\{{\arg\max}_{e\in V} F(e|S_{t-1})\}$.
For unconstrained maximization, we use the double-greedy algorithm \cite{buchbinder2015tight}, which initializes $S_1=\emptyset$ and $S_2=S_T$, where $S_T$ is the subset found in the final iteration of the greedy algorithm, and calculates $a_e=F(e|S_1)$ and $b_e=F(S_2\setminus \{e\})$ for all $e\in V$, and then adds examples for which $a_e\geq b_e$ to $S_1$ and removes examples for which $a_e< b_e$ from $S_2$ and eventually returns $S_1=S_2$.}
The complexity of the greedy algorithm is $\mathcal{O}(nk)$ to find $k$ out of $n$ examples, and can be further speed up using lazy evaluation \cite{minoux2005accelerated}.
The double-greedy applied to the subset has a complexity of $\mathcal{O}(k)$. Hence, the subset can be found efficiently. Algorithm \ref{alg} illustrates our pseudocode. 

\vspace{-2mm} \section{EXPERIMENTS} \label{sec:experiments}
 \vspace{-2mm}


In this section, we compare the zero-shot performance of training on subsets of sizes 5\%-50\% found by \method\ and those found by baselines, including C-RHO, SemDeDup, CLIP score
and Random selection. 
Moreover, we conduct an extensive ablation on the various components of \method. 

\textbf{Dataset \& Evaluation} {We use Conceptual Captions 3M and 12M \cite{sharma2018conceptual} which include 3 and 12 million image-captions pairs, respectively, and have been widely employed for benchmark evaluations in various studies focusing on contrastive language-image pre-training \cite{yang2023robust, goel2022cyclip, li2022supervision}. 
%
We evaluate all the methods on 
downstream tasks proposed by \cite{pmlr-v119-chen20j} and used in prior work for evaluating CLIP \cite{yang2023robust, goel2022cyclip, li2022supervision}. The exact list of datasets and corresponding accuracies appears in Appendix \ref{app:experiment_details}. 

\vspace{-1mm}\textbf{Training Setup} For pre-training, we use an open-source implementation of CLIP, with default ResNet-50 as the image encoder and a Transformer as the text encoder. Each experiment is run with a batch size of 512 for 30 epochs, consistent with 
\cite{yang2023robust}.\looseness=-1



\vspace{-1mm}\textbf{Baselines} The data-filtering baselines we consider are:  (1) CLIP Score \cite{gadre2023datacomp}, 
(2) C-RHO \cite{maini2023tmars}, (3) SemDeDup \cite{abbas2023semdedup}, and (4) random subsets. CLIP score
discard image-caption pairs with the smallest
similarity between their image and caption representations, obtained using a pretrained CLIP. C-RHO is an extension to RHO \cite{mindermann2022prioritized} for CLIP. It computes the similarity of paired image-caption representations using a pre-trained CLIP and compares it to the similarity obtained using a model partially trained (for 5 epochs) on the full data. Then, image-captions pairs with the smallest difference between these similarities are discarded. SemDeDup clusters the image representations of examples and then discards examples from each cluster that are most similar to each other. Due to computational constraints, we only evaluate CLIPScore and \method, the two best performing methods on CC12M. \looseness=-1



\vspace{-1mm}\textbf{Zero-Shot Performance} Fig. \ref{fig:main_results} shows that, both specifically on ImageNet and across datasets, \method\ is able to outperform previous baselines. Moreover, our results demonstrate that all common data-filtering baselines, except CLIP Score, fail to extract generalizable subsets from datasets that are already filtered. This is evidenced by these methods performing worse even than random subsets. In contrast, \method\ successfully extracts subsets that can preserve the downstream generalization performance on various datasets and outperforms CLIP Score. 
Moreover, Fig. \ref{fig:main_results} shows that \method\ can discard 50\% of the data without losing any accuracy, outperforming all baselines. In fact, only 30\% of the data is needed for performance within 90\% of training on the full data. 
Table  \ref{table:main_results} shows that \method\ achieves over 2.7x and 1.4x the accuracy of CLIP Score (the next best baseline) on ImageNet and its shifted versions. Moreover, it also shows that \method\ obtains 1.5x the accuracy of CLIP Score, across 11 downstream tasks.  Table \ref{table:cc12m_results} verifies that \method\ scales to larger datasets as well, with \method's subsets, achieving over 2.5x and 1.9x the accuracy of CLIP Score subsets, on ImageNet and its shifted versions, as well as nearly 1.25x the average accuracy over 11 downstream tasks.

{\textbf{Ablation Study} 
Table \ref{table:ablation_obj} ablates over the objective and shows that  $\Fdiv$, $\Flabelsim$ and $\Fcrossreg$ are all useful practical additions to $\Ftheory$. Table \ref{table:ablation_proxy} compares the performance of \method\ and CLIP Score where the similarities are computed using a model trained on ConceptualCaptions3M rather than the open-source CLIP provided in \cite{pmlr-v139-radford21a}. These results show that \method\ can outperform prior art, regardless of choice of proxy model. The drop in performance for both CLIP Score and \method\ when compared to the subsets in Table \ref{table:main_results}, shows that using cross-modal similarities from encoders trained on more diverse and balanced data (e.g. CLIP from \cite{pmlr-v139-radford21a}) is beneficial to both CLIP Score and \method. 

\vspace{-3mm}
\section{CONCLUSION}
\vspace{-2mm} 

{We identified subsets of examples that contribute the most to contrastive language image pre-training (\mmcl). Theoretically, we characterized the most beneficial subsets with rigorous generalization guarantees for downstream zero-shot performance, as those that preserve the cross-covariance matrix of the full training data. Empirically, we compare the performance of our method to baselines 
and show that it achieves over 2.7x and 1.4x the accuracy of the next best baseline, on ImageNet and distribution shifted versions of ImageNet. Moreover, we also show \method\ achieve 1.5x the average accuracy across 11 downstream datasets. To conclude, \method\ enables data-efficient \mmcl\ pre-training on 
massive web-scale datasets.}

\subsubsection*{Acknowledgements}
This research was partially supported by the National Science Foundation CAREER Award 2146492 and Cisco Systems. \looseness=-1

\bibliography{references}

\begin{thebibliography}{10}

\bibitem{abbas2023semdedup}
Amro Abbas, Kushal Tirumala, Dániel Simig, Surya Ganguli, and Ari~S. Morcos.
\newblock Semdedup: Data-efficient learning at web-scale through semantic
  deduplication, 2023.

\bibitem{imgnetobjectnet}
Andrei Barbu, David Mayo, Julian Alverio, William Luo, Christopher Wang, Dan
  Gutfreund, Josh Tenenbaum, and Boris Katz.
\newblock Objectnet: A large-scale bias-controlled dataset for pushing the
  limits of object recognition models.
\newblock In H.~Wallach, H.~Larochelle, A.~Beygelzimer, F.~d\textquotesingle
  Alch\'{e}-Buc, E.~Fox, and R.~Garnett, editors, {\em Advances in Neural
  Information Processing Systems}, volume~32. Curran Associates, Inc., 2019.

\bibitem{buchbinder2015tight}
Niv Buchbinder, Moran Feldman, Joseph Seffi, and Roy Schwartz.
\newblock A tight linear time (1/2)-approximation for unconstrained submodular
  maximization.
\newblock {\em SIAM Journal on Computing}, 44(5):1384--1402, 2015.

\bibitem{cc12m}
Soravit Changpinyo, Piyush Sharma, Nan Ding, and Radu Soricut.
\newblock Conceptual 12m: Pushing web-scale image-text pre-training to
  recognize long-tail visual concepts, 2021.

\bibitem{pmlr-v119-chen20j}
Ting Chen, Simon Kornblith, Mohammad Norouzi, and Geoffrey Hinton.
\newblock A simple framework for contrastive learning of visual
  representations.
\newblock In Hal~Daumé III and Aarti Singh, editors, {\em Proceedings of the
  37th International Conference on Machine Learning}, volume 119 of {\em
  Proceedings of Machine Learning Research}, pages 1597--1607. PMLR, 13--18 Jul
  2020.

\bibitem{coleman2020selection}
C~Coleman, C~Yeh, S~Mussmann, B~Mirzasoleiman, P~Bailis, P~Liang, J~Leskovec,
  and M~Zaharia.
\newblock Selection via proxy: Efficient data selection for deep learning.
\newblock In {\em International Conference on Learning Representations (ICLR)},
  2020.

\bibitem{imagenet_cvpr09}
J.~Deng, W.~Dong, R.~Socher, L.-J. Li, K.~Li, and L.~Fei-Fei.
\newblock {ImageNet: A Large-Scale Hierarchical Image Database}.
\newblock In {\em CVPR09}, 2009.

\bibitem{deng2009imagenet}
Jia Deng, Wei Dong, Richard Socher, Li-Jia Li, Kai Li, and Li~Fei-Fei.
\newblock Imagenet: A large-scale hierarchical image database.
\newblock In {\em 2009 IEEE conference on computer vision and pattern
  recognition}, pages 248--255. Ieee, 2009.

\bibitem{imgneta}
Josip Djolonga, Jessica Yung, Michael Tschannen, Rob Romijnders, Lucas Beyer,
  Alexander Kolesnikov, Joan Puigcerver, Matthias Minderer, Alexander D'Amour,
  Dan Moldovan, Sylvain Gelly, Neil Houlsby, Xiaohua Zhai, and Mario Lucic.
\newblock On robustness and transferability of convolutional neural networks,
  2021.

\bibitem{gadre2023datacomp}
Samir~Yitzhak Gadre, Gabriel Ilharco, Alex Fang, Jonathan Hayase, Georgios
  Smyrnis, Thao Nguyen, Ryan Marten, Mitchell Wortsman, Dhruba Ghosh, Jieyu
  Zhang, Eyal Orgad, Rahim Entezari, Giannis Daras, Sarah Pratt, Vivek
  Ramanujan, Yonatan Bitton, Kalyani Marathe, Stephen Mussmann, Richard Vencu,
  Mehdi Cherti, Ranjay Krishna, Pang~Wei Koh, Olga Saukh, Alexander Ratner,
  Shuran Song, Hannaneh Hajishirzi, Ali Farhadi, Romain Beaumont, Sewoong Oh,
  Alex Dimakis, Jenia Jitsev, Yair Carmon, Vaishaal Shankar, and Ludwig
  Schmidt.
\newblock Datacomp: In search of the next generation of multimodal datasets,
  2023.

\bibitem{goel2022cyclip}
Shashank Goel, Hritik Bansal, Sumit Bhatia, Ryan~A. Rossi, Vishwa Vinay, and
  Aditya Grover.
\newblock Cyclip: Cyclic contrastive language-image pretraining, 2022.

\bibitem{haochen2022provable}
Jeff~Z. HaoChen, Colin Wei, Adrien Gaidon, and Tengyu Ma.
\newblock Provable guarantees for self-supervised deep learning with spectral
  contrastive loss, 2022.

\bibitem{imgnetr}
Dan Hendrycks, Steven Basart, Norman Mu, Saurav Kadavath, Frank Wang, Evan
  Dorundo, Rahul Desai, Tyler Zhu, Samyak Parajuli, Mike Guo, Dawn Song, Jacob
  Steinhardt, and Justin Gilmer.
\newblock The many faces of robustness: A critical analysis of
  out-of-distribution generalization, 2021.

\bibitem{ji2021power}
Wenlong Ji, Zhun Deng, Ryumei Nakada, James Zou, and Linjun Zhang.
\newblock The power of contrast for feature learning: A theoretical analysis,
  2021.

\bibitem{jia2021scaling}
Chao Jia, Yinfei Yang, Ye~Xia, Yi-Ting Chen, Zarana Parekh, Hieu Pham, Quoc Le,
  Yun-Hsuan Sung, Zhen Li, and Tom Duerig.
\newblock Scaling up visual and vision-language representation learning with
  noisy text supervision.
\newblock In {\em International conference on machine learning}, pages
  4904--4916. PMLR, 2021.

\bibitem{pmlr-v202-joshi23b}
Siddharth Joshi and Baharan Mirzasoleiman.
\newblock Data-efficient contrastive self-supervised learning: Most beneficial
  examples for supervised learning contribute the least.
\newblock In Andreas Krause, Emma Brunskill, Kyunghyun Cho, Barbara Engelhardt,
  Sivan Sabato, and Jonathan Scarlett, editors, {\em Proceedings of the 40th
  International Conference on Machine Learning}, volume 202 of {\em Proceedings
  of Machine Learning Research}, pages 15356--15370. PMLR, 23--29 Jul 2023.

\bibitem{li2021supervision}
Yangguang Li, Feng Liang, Lichen Zhao, Yufeng Cui, Wanli Ouyang, Jing Shao,
  Fengwei Yu, and Junjie Yan.
\newblock Supervision exists everywhere: A data efficient contrastive
  language-image pre-training paradigm.
\newblock In {\em International Conference on Learning Representations}, 2021.

\bibitem{li2022supervision}
Yangguang Li, Feng Liang, Lichen Zhao, Yufeng Cui, Wanli Ouyang, Jing Shao,
  Fengwei Yu, and Junjie Yan.
\newblock Supervision exists everywhere: A data efficient contrastive
  language-image pre-training paradigm.
\newblock In {\em International Conference on Learning Representations}, 2022.

\bibitem{maini2023tmars}
Pratyush Maini, Sachin Goyal, Zachary~C. Lipton, J.~Zico Kolter, and Aditi
  Raghunathan.
\newblock T-mars: Improving visual representations by circumventing text
  feature learning, 2023.

\bibitem{mindermann2022prioritized}
Sören Mindermann, Jan Brauner, Muhammed Razzak, Mrinank Sharma, Andreas
  Kirsch, Winnie Xu, Benedikt Höltgen, Aidan~N. Gomez, Adrien Morisot,
  Sebastian Farquhar, and Yarin Gal.
\newblock Prioritized training on points that are learnable, worth learning,
  and not yet learnt, 2022.

\bibitem{minoux2005accelerated}
Michel Minoux.
\newblock Accelerated greedy algorithms for maximizing submodular set
  functions.
\newblock In {\em Optimization Techniques: Proceedings of the 8th IFIP
  Conference on Optimization Techniques W{\"u}rzburg, September 5--9, 1977},
  pages 234--243. Springer, 2005.

\bibitem{mirzasoleiman2016fast}
Baharan Mirzasoleiman, Ashwinkumar Badanidiyuru, and Amin Karbasi.
\newblock Fast constrained submodular maximization: Personalized data
  summarization.
\newblock In {\em International Conference on Machine Learning}, pages
  1358--1367. PMLR, 2016.

\bibitem{mu2022slip}
Norman Mu, Alexander Kirillov, David Wagner, and Saining Xie.
\newblock Slip: Self-supervision meets language-image pre-training.
\newblock In {\em European Conference on Computer Vision}, pages 529--544.
  Springer, 2022.

\bibitem{pmlr-v206-nakada23a}
Ryumei Nakada, Halil~Ibrahim Gulluk, Zhun Deng, Wenlong Ji, James Zou, and
  Linjun Zhang.
\newblock Understanding multimodal contrastive learning and incorporating
  unpaired data.
\newblock In Francisco Ruiz, Jennifer Dy, and Jan-Willem van~de Meent, editors,
  {\em Proceedings of The 26th International Conference on Artificial
  Intelligence and Statistics}, volume 206 of {\em Proceedings of Machine
  Learning Research}, pages 4348--4380. PMLR, 25--27 Apr 2023.

\bibitem{paul2023deep}
Mansheej Paul, Surya Ganguli, and Gintare~Karolina Dziugaite.
\newblock Deep learning on a data diet: Finding important examples early in
  training, 2023.

\bibitem{pooladzandi2022adaptive}
Omead Pooladzandi, David Davini, and Baharan Mirzasoleiman.
\newblock Adaptive second order coresets for data-efficient machine learning,
  2022.

\bibitem{pmlr-v139-radford21a}
Alec Radford, Jong~Wook Kim, Chris Hallacy, Aditya Ramesh, Gabriel Goh,
  Sandhini Agarwal, Girish Sastry, Amanda Askell, Pamela Mishkin, Jack Clark,
  Gretchen Krueger, and Ilya Sutskever.
\newblock Learning transferable visual models from natural language
  supervision.
\newblock In Marina Meila and Tong Zhang, editors, {\em Proceedings of the 38th
  International Conference on Machine Learning}, volume 139 of {\em Proceedings
  of Machine Learning Research}, pages 8748--8763. PMLR, 18--24 Jul 2021.

\bibitem{imgnetv2}
Benjamin Recht, Rebecca Roelofs, Ludwig Schmidt, and Vaishaal Shankar.
\newblock Do imagenet classifiers generalize to imagenet?, 2019.

\bibitem{sharma2018conceptual}
Piyush Sharma, Nan Ding, Sebastian Goodman, and Radu Soricut.
\newblock Conceptual captions: A cleaned, hypernymed, image alt-text dataset
  for automatic image captioning.
\newblock In {\em Proceedings of ACL}, 2018.

\bibitem{imgnetsketch}
Haohan Wang, Songwei Ge, Eric~P. Xing, and Zachary~C. Lipton.
\newblock Learning robust global representations by penalizing local predictive
  power, 2019.

\bibitem{pmlr-v202-xue23d}
Yihao Xue, Siddharth Joshi, Eric Gan, Pin-Yu Chen, and Baharan Mirzasoleiman.
\newblock Which features are learnt by contrastive learning? {O}n the role of
  simplicity bias in class collapse and feature suppression.
\newblock In Andreas Krause, Emma Brunskill, Kyunghyun Cho, Barbara Engelhardt,
  Sivan Sabato, and Jonathan Scarlett, editors, {\em Proceedings of the 40th
  International Conference on Machine Learning}, volume 202 of {\em Proceedings
  of Machine Learning Research}, pages 38938--38970. PMLR, 23--29 Jul 2023.

\bibitem{yang2023robust}
Wenhan Yang and Baharan Mirzasoleiman.
\newblock Robust contrastive language-image pretraining against adversarial
  attacks, 2023.

\bibitem{pmlr-v202-yang23g}
Yu~Yang, Hao Kang, and Baharan Mirzasoleiman.
\newblock Towards sustainable learning: Coresets for data-efficient deep
  learning.
\newblock In {\em Proceedings of the 40th International Conference on Machine
  Learning}, volume 202 of {\em Proceedings of Machine Learning Research},
  pages 39314--39330. PMLR, 23--29 Jul 2023.

\bibitem{devil}
Haichao Yu, Yu~Tian, Sateesh Kumar, Linjie Yang, and Heng Wang.
\newblock The devil is in the details: A deep dive into the rabbit hole of data
  filtering, 2023.

\bibitem{pmlr-v202-zhang23an}
Qi~Zhang, Yifei Wang, and Yisen Wang.
\newblock On the generalization of multi-modal contrastive learning.
\newblock In Andreas Krause, Emma Brunskill, Kyunghyun Cho, Barbara Engelhardt,
  Sivan Sabato, and Jonathan Scarlett, editors, {\em Proceedings of the 40th
  International Conference on Machine Learning}, volume 202 of {\em Proceedings
  of Machine Learning Research}, pages 41677--41693. PMLR, 23--29 Jul 2023.

\end{thebibliography}
\bibliographystyle{plain}

\clearpage
\section*{Checklist}

 \begin{enumerate}

 \item For all models and algorithms presented, check if you include:
 \begin{enumerate}
   \item A clear description of the mathematical setting, assumptions, algorithm, and/or model. \textbf{Yes}
   \item An analysis of the properties and complexity (time, space, sample size) of any algorithm. \textbf{Yes}
   \item (Optional) Anonymized source code, with specification of all dependencies, including external libraries. \textbf{Yes}
 \end{enumerate}

 \item For any theoretical claim, check if you include:
 \begin{enumerate}
   \item Statements of the full set of assumptions of all theoretical results. \textbf{Yes}
   \item Complete proofs of all theoretical results. \textbf{Yes}
   \item Clear explanations of any assumptions. \textbf{Yes}     
 \end{enumerate}

 \item For all figures and tables that present empirical results, check if you include:
 \begin{enumerate}
   \item The code, data, and instructions needed to reproduce the main experimental results (either in the supplemental material or as a URL). \textbf{Yes}
   \item All the training details (e.g., data splits, hyperparameters, how they were chosen). \textbf{Yes}
         \item A clear definition of the specific measure or statistics and error bars (e.g., with respect to the random seed after running experiments multiple times). \textbf{N/A}
         \item A description of the computing infrastructure used. (e.g., type of GPUs, internal cluster, or cloud provider). \textbf{Yes}
 \end{enumerate}

 \item If you are using existing assets (e.g., code, data, models) or curating/releasing new assets, check if you include:
 \begin{enumerate}
   \item Citations of the creator If your work uses existing assets. \textbf{N/A}
   \item The license information of the assets, if applicable. \textbf{N/A}
   \item New assets either in the supplemental material or as a URL, if applicable. \textbf{N/A}
   \item Information about consent from data providers/curators. \textbf{N/A}
   \item Discussion of sensible content if applicable, e.g., personally identifiable information or offensive content. \textbf{N/A}
 \end{enumerate}

 \item If you used crowdsourcing or conducted research with human subjects, check if you include:
 \begin{enumerate}
   \item The full text of instructions given to participants and screenshots. \textbf{N/A}
   \item Descriptions of potential participant risks, with links to Institutional Review Board (IRB) approvals if applicable. \textbf{N/A}
   \item The estimated hourly wage paid to participants and the total amount spent on participant compensation. \textbf{N/A}
 \end{enumerate}

 \end{enumerate}

\onecolumn
\appendix

\section{Experimental Details}\label{app:experiment_details}





\subsection{Accuracy on downstream tasks}
\begin{small}
\begin{table}[H]
\caption{5\% \method\ subset selected from CC3M}\label{table:5pct_downstream_results}
\centering
\begin{tabular}{c|lllll}
\textbf{Datasets}  & \textbf{Random}& \textbf{C-RHO} & \textbf{SemDeDup} & \textbf{CLIP Score }& \textbf{\method} \\ \hline
Caltech101         & 9.71\%& 3.24\%& 5.52\%& 11.22\%&16.93\%\\ 
DTD                & 3.72\%& 2.23\%& 2.61\%& 3.78\%&4.10\%\\ 
Food101            & 1.90\%& 1.05\%& 1.15\%& 2.58\%&3.24\%\\ 
ImageNet           & 1.27\%& 0.42\%& 0.85\%& 1.65\%&4.46\%\\
STL10              & 17.57\%& 16.45\%& 22.44\%& 15.10\%&22.31\%\\ 
SUN397             & 3.82\%& 0.79\%& 1.89\%& 2.70\%&5.33\%\\ 
ImageNet-Sketch    & 0.23\%& 0.24\%& 0.28\%& 0.68\%&0.84\%\\ 
ImageNet-V2        & 1.18\%& 0.42\%& 0.75\%& 1.53\%&3.76\%\\ 
ImageNet-A         & 1.17\%& 0.76\%& 1.15\%& 1.44\%&1.55\%\\ 
ImageNet-R         & 2.32\%& 1.13\% & 1.37\%& 4.22\%&5.52\%\\ 
ObjectNet          & 0.60\%& 0.41\%& 0.53\%& 0.91\%&1.07\%\\ 
\end{tabular}
\end{table}

\begin{table}[H]
\caption{10\% \method\ subset selected from CC3M}\label{table:10pct_downstream_results}
\centering
\begin{tabular}{c|lllll}
\textbf{Datasets}  & \textbf{Random}& \textbf{C-RHO} & \textbf{SemDeDup} & \textbf{CLIP Score }& \textbf{\method} \\ \hline
Caltech101         & 18.67\%& 13.82\%& 16.60\%& 36.45\%&32.44\%\\ 
DTD                & 3.72\%& 4.31\%& 2.02\%& 8.94\%&9.31\%\\ 
Food101            & 3.40\%& 1.68\%& 0.99\%& 5.03\%&5.36\%\\ 
ImageNet           & 3.95\%& 1.89\%& 2.60\%& 6.48\%&11.33\%\\
STL10              & 26.10\%& 19.56\%& 23.93\%& 22.96\%&35.01\%\\ 
SUN397             & 9.92\%& 3.01\%& 5.49\%& 11.18\%&15.78\%\\ 
ImageNet-Sketch    & 1.13\%& 0.77\%& 0.64\%& 2.76\%&3.89\%\\ 
ImageNet-V2        & 3.66\%& 1.66\%& 2.55\%& 5.00\%&9.04\%\\ 
ImageNet-A         & 1.37\%& 1.28\%& 1.45\%& 1.69\%&2.07\%\\ 
ImageNet-R         & 4.87\%& 3.21\%& 3.65\%& 11.02\%&12.71\%\\ 
ObjectNet          & 1.11\%& 0.86\%& 1.07\%& 1.74\%&2.12\%\\ 
\end{tabular}
\end{table}

\begin{table}[H]
\caption{20\% \method\ subset Sizes Per Dataset Accuracies}\label{table:20pct_downstream_results}
\centering
\begin{tabular}{c|lllll}
\textbf{Datasets}  & \textbf{Random}& \textbf{C-RHO} & \textbf{SemDeDup} & \textbf{CLIP Score }& \textbf{\method} \\ \hline
Caltech101         & 34.30\%& 33.36\%& 30.70\%& 40.28\%&45.83\%\\ 
DTD                & 7.93\%& 7.71\%& 4.41\%& 12.39\%&11.49\%\\ 
Food101            & 5.83\%& 4.05\%& 4.22\%& 9.51\%&9.49\%\\ 
ImageNet           & 7.99\%& 7.39\%& 6.25\%& 12.79\%&15.86\%\\
STL10              & 32.72\%& 32.31\%& 28.07\%& 36.30\%&42.24\%\\ 
SUN397             & 17.81\%& 12.71\%& 13.93\%& 22.25\%&24.89\%\\ 
ImageNet-Sketch    & 2.48\%& 3.69\%& 1.82\%& 6.94\%&7.77\%\\ 
ImageNet-V2        & 7.12\%& 6.34\%& 5.23\%& 10.72\%&13.38\%\\ 
ImageNet-A         & 1.91\%& 2.17\%& 1.83\%& 2.43\%&2.17\%\\ 
ImageNet-R         & 9.14\%& 10.26\%& 7.32\%& 17.55\%&19.32\%\\ 
ObjectNet          & 2.06\%& 2.45\%& 2.05\%& 3.40\%&3.54\%\\ 
\end{tabular}
\end{table}

\begin{table}[H]
\caption{30\% \method\ subset selected from CC3M}\label{table:30pct_downstream_results}
\centering
\begin{tabular}{c|lllll}
\textbf{Datasets}  &    \textbf{Random} &\textbf{C-RHO}&\textbf{SemDeDup}&\textbf{CLIP Score }& \textbf{\method} \\ \hline
Caltech101         &    38.18\%&42.47\%&36.12\%&47.25\%&48.17\%\\ 
DTD                &    8.62\%&10.90\%&9.15\%&14.63\%&12.82\%\\ 
Food101            &    6.85&5.86\%&5.76\%&12.65\%&12.54\%\\ 
ImageNet           &    10.21\%&11.01\%&9.32\%&18.54\%&19.12\%\\
STL10              &    39.24\%&39.85\%&39.56\%&47.83\%&45.95\%\\ 
SUN397             &    21.09\%&19.14\%&19.19\%&31.41\%&37.04\%\\ 
ImageNet-Sketch    &    4.28\%&5.94\%&3.12\%&10.19\%&10.35\%\\ 
ImageNet-V2        &    8.94\%&9.20\%&7.09\%&16.06\%&16.09\%\\ 
ImageNet-A         &    2.12\%&2.67\%&2.51\%&2.47\%&3.35\%\\ 
ImageNet-R         &    12.21\%&14.32\%&10.56\%&22.39\%&23.50\%\\ 
ObjectNet          &    2.57\%&3.20\%&2.56\%&4.91\%&4.98\%\\ 
\end{tabular}
\end{table}

\begin{table}[H]
\caption{40\% \method\ subset selected from CC3M}\label{table:40pct_downstream_results}
\centering
\begin{tabular}{c|lllll}
\textbf{Datasets}  & \textbf{Random}& \textbf{C-RHO} & \textbf{SemDeDup} & \textbf{CLIP Score }& \textbf{\method} \\ \hline
Caltech101         & 39.31\%& 45.47\%& 40.67\%& 51.73\%&52.08\%\\ 
DTD                & 12.61\%& 11.12\%& 9.57\%& 12.93\%&12.39\%\\ 
Food101            & 8.51\%& 7.08\%& 8.01\%& 12.68\%&12.61\%\\ 
ImageNet           & 11.99\%& 13.56\%& 11.34\%& 17.48\%&18.47\%\\
STL10              & 42.14\%& 43.01\%& 40.27\%& 46.11\%&45.85\%\\ 
SUN397             & 24.41\%& 23.39\%& 24.08\%& 30.63\%&31.49\%\\ 
ImageNet-Sketch    & 5.05\%& 6.56\%& 4.60\%& 9.84\%&10.17\%\\ 
ImageNet-V2        & 9.98\%& 11.86\%& 9.34\%& 14.57\%&15.25\%\\ 
ImageNet-A         & 2.51\%& 2.91\%& 2.64\%& 2.97\%&2.80\%\\ 
ImageNet-R         & 14.09\%& 16.90\%& 13.36\%& 22.43\%&22.10\%\\ 
ObjectNet          & 3.16\%& 4.05\%& 3.32\%& 4.79\%&4.84\%\\ 
\end{tabular}
\end{table}

\begin{table}[H]
\caption{50\% \method\ subset selected from CC3M}\label{table:50pct_downstream_results}
\centering
\begin{tabular}{c|lllll}
\textbf{Datasets}  & \textbf{Random}& \textbf{C-RHO} & \textbf{SemDeDup} & \textbf{CLIP Score }& \textbf{\method} \\ \hline
Caltech101         & 41.96\%& 46.80\%& 44.06\%& 47.25\%&48.17\%\\ 
DTD                & 9.20\%& 10.96\%& 8.51\%& 14.63\%&12.82\%\\ 
Food101            & 8.44\%& 8.42\%& 8.39\%& 12.65\%&12.54\%\\ 
ImageNet           & 13.45\%& 14.56\%& 13.28\%& 18.54\%&19.12\%\\
STL10              & 45.29\%& 46.40\%& 44.46\%& 47.83\%&45.95\%\\ 
SUN397             & 24.41\%& 25.78\%& 25.88\%& 31.41\%&37.04\%\\ 
ImageNet-Sketch    & 6.45\%& 7.96\%& 5.65\%& 10.19\%&10.35\%\\ 
ImageNet-V2        & 11.55\%& 12.11\%& 11.08\%& 16.06\%&16.09\%\\ 
ImageNet-A         & 2.37\%& 3.05\%& 3.03\%& 2.47\%&3.35\%\\ 
ImageNet-R         & 15.63\%& 17.40\%& 14.85\%& 22.39\%&23.50\%\\ 
ObjectNet          & 3.47\%& 4.36\%& 3.45\%& 4.91\%&4.98\%\\ 
\end{tabular}
\end{table}

\end{small}


\subsection{Additional Training Details}

The experiments were conducted using NVIDIA A100s and NVIDIA RTX A6000 GPUs.
\clearpage
\subsection{Visualization of CC12M Results}

\begin{figure*}[h]
    \centering
    \begin{subfigure}[b]{0.3\textwidth}
        \centering
        \includegraphics[width=\linewidth]{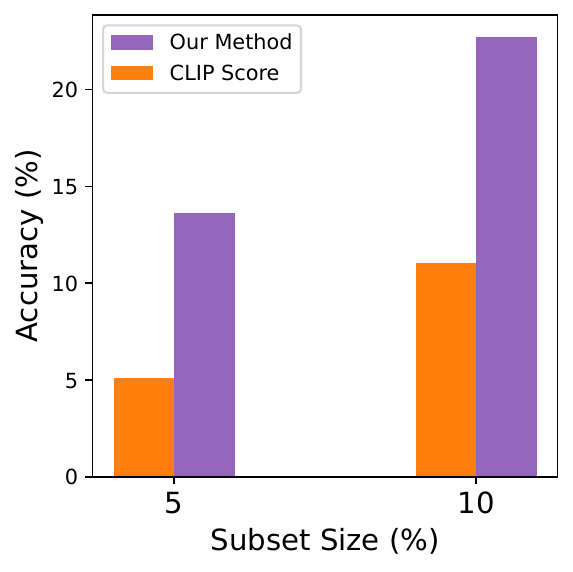}
        \caption{ImageNet}
    \end{subfigure}
    \hfill
    \begin{subfigure}[b]{0.3\textwidth}
        \centering
        \includegraphics[width=\linewidth]{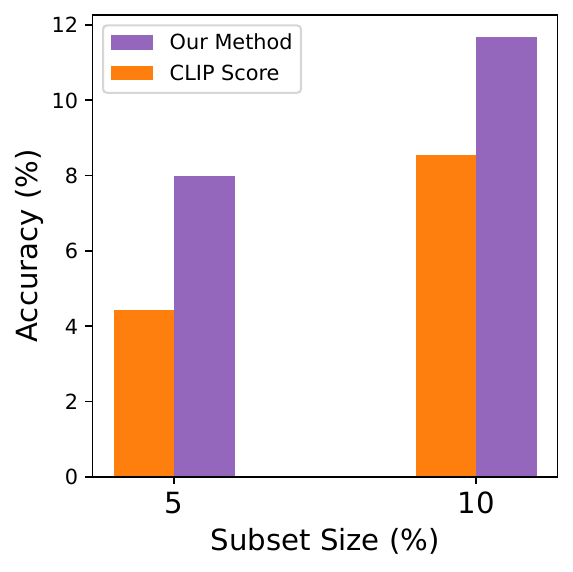}
        \caption{ImageNet Dist. Shift}
    \end{subfigure}
    \hfill
    \begin{subfigure}[b]{0.3\textwidth}
        \centering
        \includegraphics[width=\linewidth]{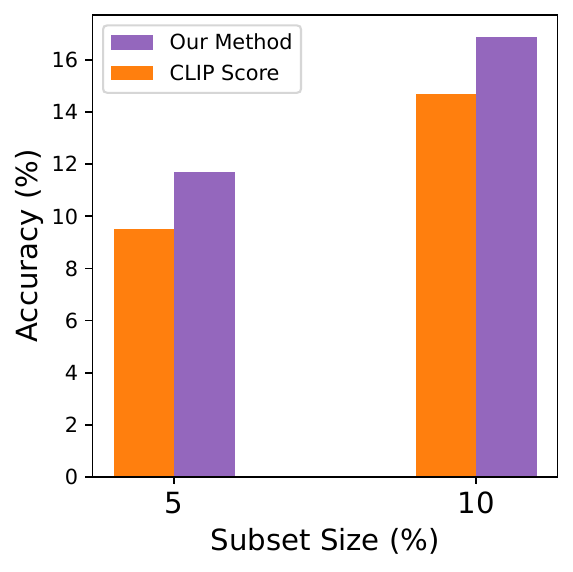}
        \caption{Avg. over 11 Datasets}
    \end{subfigure}
    \caption{Performance across subset of sizes 5\% and 10\% from CC12M}
    \label{fig:cc12m_results}
\end{figure*}

\end{document}